\documentclass[conference]{IEEEtran}
\IEEEoverridecommandlockouts
\usepackage{cite}
\usepackage{amsmath,amssymb,amsfonts}
\usepackage{algorithm}
\usepackage{algorithmic}
\usepackage{multirow}
\usepackage{graphicx}
\usepackage{textcomp}
\usepackage{amsmath}
\usepackage{xcolor}
\def\BibTeX{{\rm B\kern-.05em{\sc i\kern-.025em b}\kern-.08em
    T\kern-.1667em\lower.7ex\hbox{E}\kern-.125emX}}
\begin{document}

\title{Thief, Beware of What Get You There: Towards Understanding Model Extraction Attack
}

\author{\IEEEauthorblockN{Xinyi Zhang}
\IEEEauthorblockA{\textit{School of Computer Science and Engineering} \\
\textit{Nanyang Technological University}\\
Singapore \\
zh0031yi@e.ntu.edu.sg}
\and
\IEEEauthorblockN{Chengfang Fang}
\IEEEauthorblockA{\textit{Singapore Research Center} \\
\textit{Huawei International Pte. Ltd.}\\
Singapore \\
fang.chengfang@huawei.com}
\and
\IEEEauthorblockN{Jie Shi}
\IEEEauthorblockA{\textit{Singapore Research Center} \\
\textit{Huawei International Pte. Ltd.}\\
Singapore \\
shi.jie1@huawei.com}
}

\maketitle

\begin{abstract}
Model extraction increasingly attracts research attentions as keeping commercial AI models private can retain a competitive advantage. In some scenarios, AI models are trained proprietarily, where neither pre-trained models nor sufficient in-distribution data is publicly available. Model extraction attacks against these models are typically more devastating. Therefore, in this paper, we empirically investigate the behaviors of model extraction under such scenarios. We find the effectiveness of existing techniques significantly affected by the absence of pre-trained models. In addition, the impacts of the attacker's hyperparameters, e.g. model architecture and optimizer, as well as the utilities of information retrieved from queries, are counterintuitive. We provide some insights on explaining the possible causes of these phenomena. With these observations, we formulate model extraction attacks into an adaptive framework that captures these factors with deep reinforcement learning. Experiments show that the proposed framework can be used to improve existing techniques, and show that model extraction is still possible in such strict scenarios. Our research can help system designers to construct better defense strategies based on their scenarios.
\end{abstract}

\section{Introduction}
With Artificial Intelligence capabilities increasingly embedded in enterprise tools, products, and services, companies who adopted AI in their business have grown and expanded steeply. AI becomes one of the important factors for them to gain competitive advantages over other players. Thus, to retain their advantage, protecting the intelligent properties from competitors is an important topic. 

To maximize the advantage, AI capability is typically deployed as a service that is available to the public, their customers, or their applications. There are rising concerns on model extraction attacks through such services ever since \cite{tramer2016stealing}, which showed that it is possible to extract a similar model from the available API. A series of attacks and defense mechanisms with different assumptions are proposed, and the concern is extended to the more complicated models like ResNet \cite{he2016deep} and BERT \cite{devlin2018bert}.

Most of the current researches on model extraction and defense are empirical, with only few exceptions such as \cite{jagielski2019high}. Many of the empirical model extractions are conducted with public datasets and pre-trained models, and thus assume the attacker who has access to a similarly trained model and in-distribution datasets. Despite this being a straightforward approach for an attacker who tries to extract these models, there are scenarios where the models are trained with in-house data, and neither the data distribution nor  pre-trained model is available to the attacker. In this paper, we will discuss this less addressed scenario. Nevertheless, separating the contributing factors of model extraction, in other words, studying the effectiveness of model extraction without a pre-trained model or in-distribution dataset, will help us understand the fundamentals of model extraction.

Intuitively, there are two important factors that could affect the effectiveness of model extraction, namely, attacker model hyperparameters and information retrieved from the victim through queries. We firstly conduct experiments to analyze the effect of model structure, and show that the optimal structure cannot be simply determined by the victim model structure, and thus shall be configured adaptively. 

Secondly, we conduct experiments to show that, the learning setting in model extraction might be counter-intuitive compared to normal model training. For example, due to the query distribution, the Adam \cite{kingma2017adam} optimizer might lead to a significantly worse result compared to the SGD. A learning rate and training epoch number that might lead to overfitting in model training might be favorable for model extraction.

Lastly, we examine the effect of query set. Specifically, we study the extraction effect in three typical scenarios: when the adversary does not have data of specific labels, when the amount of in-distribution data is limited, and when the adversary does not have any in-distribution data. The experiment shows that extraction is still possible even if the malicious party does not possess the corresponding data.

Since the adversary will always try to expand the query set until the query budget is met, the extraction effect also directly depends on the query data generation strategy. In view of the aforementioned constraints and phenomena, we take one more step towards understanding model extraction attack by proposing a heuristic extraction framework utilizing deep reinforcement learning, with which we learn the optimal query strategy. Experiments show that one example implementation of our framework can outperform the typical FGSM \cite{goodfellow2014explaining} method, and can extract a reasonable model even without any in-distribution data and within a limited budget, which thus shall be considered and utilized by security architects who intend to rely on the secrecy of data to achieve  model confidentiality.

The rest of the paper is organized as follow: in section \ref{sec:relatedwork} we briefly summarize existing works that are related to this research; in section \ref{sec:problem} we introduce the notation and problem formulation of this paper; in section \ref{sec:analysis} we present our observations and analysis; in section \ref{sec:experiment}, we propose and validate the model extraction framework.

\section{Related Work}
\label{sec:relatedwork}
\subsection{Model Extraction} 
Model extraction (ME) typically refers to extracting non-public information, such as functionality or parameters, from a black-box Machine Learning model. In the context of AI Security, it is also known as Model Stealing Attack due to its nature of being unauthorized. Earlier works in model stealing attack focus on simple ML models like SVM or decision tree \cite{tramer2016stealing}. As Deep Learning gains popularity, recent works \cite{orekondy2019knockoff,wang2020neural,krishna2020thieves} started to explore complex deep neural networks. These works present the effectiveness of their attacking methodologies in Computer Vision (Image Classifier) or Nature Language Processing settings. Most of these works rely on open-source pre-trained weight, high-quality data, or sophisticated models. However, such given prior knowledge could be the bottleneck of attacks in practical situations. In our work, we study model extraction attack in stricter situations and point out critical but lightly mentioned factors. 

%in order to clear pitfalls for researchers entering AI Security field with a fast-growing number.

\subsection{Knowledge Distillation}
A closely related research topic is knowledge distillation (KD), which refers to transferring the functionality from one model to another. It typically involves a teacher model, which is structurally complex, and a student model, which is simpler and easier to deploy. It was firstly proposed by \cite{hinton2015distilling} and has been constantly studied since for better efficiency and performance \cite{nayak2019zero,li2020few}. We argue that essentially, knowledge distillation is a special case of a functionality-targeted Model Extraction problem. In opposite to the black-box setting in Model Extraction, knowledge distillation can leverage the knowledge of the teacher model, its training data, the activation \cite{zagoruyko2016paying,lopes2017data}, and structural information \cite{8100237} for various improvements. In this paper, we combine some phenomena that occurred in our experiments with observations \cite{yang2018knowledge, cho2019efficacy} made in KD, to better understands the effectiveness of ME.

\subsection{Adversarial Example}
With a human-imperceptible synthesized noise \cite{carlini2017towards} added to the input, adversarial examples (AE) are known to be capable of greatly affecting the output of neural networks. Due to its nature of manipulating the model output, AE is also used in KD \cite{heo2019knowledge} and ME \cite{papernot2017practical} as a way to find decision-boundary. The explanation of AE still remains unclear, but multiple theories \cite{gilmer2018adversarial,shamir2019simple,ilyas2019adversarial} have been proposed. Reference \cite{ilyas2019adversarial} interpreted the AE noise as non-robust features, which we adopted in our method to enrich the information retrieved from the victim and push the distribution of the sampled data to a desired state.

\subsection{Active Learning}
Active learning (AL) is a technique where data is adaptively selected and trained \cite{settles2009active,zhou2017fine} based on the state of the evolving model. The purpose of AL is to save data labeling effort, which, in the context of ME, is the interaction with the target black-box model. In this work, we use deep reinforcement learning (DRL), which has been proven to able to outperform human-level control \cite{mnih2015humanlevel} in many systems, in section \ref{sec:experiment} for our proposed example algorithm to conduct active learning.

\section{Problem Formulation}
\label{sec:problem}
%\subsection{Notations}
In this paper, we mainly follow the notation defined in \cite{orekondy2019knockoff} and consider the setting where an attacker wants to train a model $F_a$ by querying a victim model $F_v: \mathcal{X} \xrightarrow{} \mathcal{Y}$, which is trained on dataset $D_v$. The attacker is provided with a data pool $D_a$ containing labeled, non-labeled, or no data. The attacker has a budget $B$, that is, the number of times he can query the victim model. The set of queries he constructed is referred to as the transfer set $D_t$.

We consider model extraction problem as a ``Task Accuracy Extraction'' defined in \cite{jagielski2019high}. That is, we analyze the effect of $D_v$, model structure, learning parameters that affect the accuracy over a task distribution $D_{task}$ over $\mathcal{X} \times \mathcal{Y}$, i.e.: 
\begin{equation}
    \Pr_{(x,y)\in D_{task}} [argmax(F_a(x))=y]
\end{equation}%

We study the following two factors which typically affect the effectiveness of model extraction: 

\subsubsection{Query Utility} As some of the existing works pointed out, the efficiency of model extraction is affected by the utilities of information retrieved from queries, including: (1) whether the queries are in-distribution, i.e. whether they are in the set of $D_T$, or even $D_v$; (2) whether the queries are balanced in terms of each label in $\mathcal{Y}$; (3) whether the queries sufficiently explore the input space $\mathcal{X}$.

\subsubsection{Hyperparameters} We analyze the effect of model hyperparameters by comparing the learning process of the following three cases: where the attacker uses a more complex structure, the same structure, and a simpler structure.  We also analyze the effect of learning hyperparameters such as optimizer.

\begin{table*}[!!htbp]
\caption{Knockoff Nets Performance with Varying Attacker Model Pre-trained Weights}\label{tab:caltech}
\begin{center}
\begin{tabular}{|c|c|c|c|c|c|c|c|}
\hline
$\boldsymbol{D_v}$ & $\boldsymbol{F_v}$ & $\boldsymbol{F_v\ ACC}^{\mathrm{a}}$~\textbf{(\%)} & $\boldsymbol{F_a}$ & $\boldsymbol{D_a}$ & $\boldsymbol{F_a}$~\textbf{Pre-trained} & $\boldsymbol{B}^{\mathrm{b}}$ & $\boldsymbol{F_a\ ACC}^{\mathrm{a}}$~\textbf{(\%)} \\
\hline
\multirow{6}{*}{Caltech256} & \multirow{4}{*}{ResNet-34} & \multirow{4}{*}{78.4} & ResNet-34 & \multirow{4}{*}{Caltech256} & ImageNet & \multirow{4}{*}{$10^{4}$}  & 78.1     \\ \cline{4-4} \cline{6-6} \cline{8-8}
& & & ResNet-34 & & \multirow{8}{*}{N/A$^{\mathrm{c}}$} & & 20.1 \\ \cline{4-4} \cline{8-8}
& & & ResNet-50 & & & & 21.1 \\ \cline{4-4} \cline{8-8}
& & & ResNet-18 & & & & 21.8 \\ \cline{2-5} \cline{7-8}
& \multirow{2}{*}{ResNet-50} & \multirow{2}{*}{85.2} & \multirow{2}{*}{ResNet-18} & Caltech256 & & $10^{4}$ & 22.4 \\ \cline{5-5} \cline{7-8}
& & & & ImageNet & & $10^{5}$ & 29.7 \\ \cline{1-5} \cline{7-8}
\multirow{3}{*}{ImageNet} & \multirow{3}{*}{ResNet-50} & \multirow{3}{*}{84.5} & ResNet-18 &\multirow{3}{*}{ImageNet} & &\multirow{2}{*}{$10^{5}$} & 23.6    \\ \cline{4-4} \cline{8-8}
& & & ResNet-50 & & & & 18.1    \\ \cline{4-4} \cline{7-8}
& & & ResNet-50 & & & $10^{6}$ & 77.8    \\
\hline
\multicolumn{8}{l}{$^{\mathrm{a}}$$ACC$: Observed best top-1 test accuracy.} \\
\multicolumn{8}{l}{$^{\mathrm{b}}$Random sampling as in the original paper applied.} \\
\multicolumn{8}{l}{$^{\mathrm{c}}$N/A: The attacker model is not trained on any dataset before the starting \ of the extraction process.} \\
\end{tabular}
\end{center}
\end{table*}

\section{Observation and Analysis}
\label{sec:analysis}
Most of the existing works study model extraction with image classification tasks and therefore assume the attacker starts with a pre-trained model (e.g. ResNet-34 on ImageNet \cite{imagenet_cvpr09}). While this is reasonable when attackers try to extract models with common tasks, it might not be the case for some of the commercial critical tasks, where even training data are confidential properties. In fact, model extraction attacks are more probable and severe for these cases. Nevertheless, for better comparison with existing works, we conduct experiments with image classification tasks. However, for most of our analysis, we do not assume the attacker starts with a pre-trained model, nor does he have plenty in-distribution data. 

We study the effects of the aforementioned factors by extending the experiments in Knockoff Nets \cite{orekondy2019knockoff}. All tests are conducted on the standard test dataset or using the same train-test split as in the original paper. The victim model is assumed to return the softmaxed classification probability (soft label) of the queried input. We organize our observations into the following three categories:
\subsection{Pre-trained Weight}

In this section, we discuss the effect of $F_a$ pre-trained weight and the role it plays in the whole extraction process. We start with replicating Knockoff Nets random sampling on Caltech-256 \cite{griffin_caltech-256_2007}. Then, instead of using ResNet-34 pre-trained on ImageNet, we apply Xavier \cite{Glorot10understandingthe} initialization on three types of attacker model structures and repeated the experiment. To avoid coincidence and rule out factors like incapable victim or inadequate samples, we also change $F_v$ for more accurate models, vary $B$, and extend experiments to ImageNet.

The result (Table \ref{tab:caltech}) shows that the performance of $F_a$ degenerates significantly without pre-trained weight, even when it has matching network structure with $F_v$ and $D_v$ as query set. A related result is also found in \cite{asokan2020extraction}, where they showed that $F_a$ performance decreases when $F_v$ is not finetuned from the same pre-trained model.

Based on the above-mentioned observation, we find that when the victim and attacker both start with models pre-trained on the same dataset, the importance of the pre-trained weight is more than providing just a sophisticated feature extractor, but rather significant prior knowledge of $F_v$ and make the whole process similar to fine-tuned transfer learning. Without such pre-trained weight, the attacker needs a much higher budget to obtain a comparable result.

\subsection{Attacker Hyperparameters}
\label{subsec:struc}
In this paper, we look into the behaviors of two important factors, namely attacker model structure and optimizer, in model extraction. Our Experiments show that their behaviors differ from those under typical machine learning settings.
\subsubsection{Attacker Model Structure}
Previous works  \cite{orekondy2019knockoff,krishna2020thieves} studied the effect of structure in the settings of $F_a$ pre-trained weight. Conclusions were made that:
\begin{itemize}
    \item Given fixed $F_v$ structure, the more complex $F_a$ is, the better the extraction result is.
    \item Given fixed $F_a$ structure, the extraction result is the best when $F_a = F_v$.
\end{itemize}
In contrast, in this section, we study the effect $F_a$ structure without pre-trained weight, in order to validate the above-mentioned two statements in a broader context. 

\begin{table}[b]
\caption{Results of Model Extraction on FashionMNIST Using White Noise With Varying Attacker and Victim Model Structure}\label{tab:structure}
\begin{center}
\begin{tabular}{|c|c|c|c|}
\hline
$\boldsymbol{F_v}$ & $\boldsymbol{F_v\ ACC}^{\mathrm{a}}$~\textbf{(\%)} & $\boldsymbol{F_a}$ & $\boldsymbol{F_a\ ACC}^{\mathrm{a}}$~\textbf{(\%)}\\
\hline
\multirow{3}{*}{MLP} & \multirow{3}{*}{88.0} & MLP & 40.9 \\ \cline{3-4}
 & & LeNet & 28.1 \\ \cline{3-4}
 & & AlexNet & 24.1 \\ \cline{1-4}
\multirow{3}{*}{LeNet} & \multirow{3}{*}{90.7} & MLP & 23.6 \\ \cline{3-4}
 && LeNet & 62.6 \\ \cline{3-4}
 && AlexNet & 25.1 \\ \cline{1-4}
\multirow{3}{*}{AlexNet} & \multirow{3}{*}{91.1} & MLP & 11.5 \\\cline{3-4}
 && LeNet & 15.8 \\\cline{3-4}
 && AlexNet & 23.7 \\
\hline
\multicolumn{4}{l}{$^{\mathrm{a}}$$ACC$: Observed best top-1 test accuracy.} \\
\end{tabular}
\end{center}
\end{table}

We conduct experiments on FashionMNIST \cite{xiao2017fashion} dataset with three different models: 2-layer MLP, LeNet \cite{lecun1998gradient}, and resized AlexNet \cite{NIPS2012_c399862d}, of $10^5$, $4*10^5$, and $4*10^6$ trainable parameters respectively. The experiment result (Table \ref{tab:structure}) indicates that with no pre-trained weight applied, $F_a$ is neither the more complex the better nor performs the best when $F_v$ has the same structure. Similar phenomena can also be observed in Table~\ref{tab:caltech}, where ResNet-50 underperforms ResNet-18 with the same budget.

We believe that in previous works' settings, pre-trained complex models reveal more information about the original feature space that $F_v$ was trained on, which is the major factor that distinguished their performance from the simple ones. This could be an alternative explanation to the fact that, in the original paper, when testing Knockoff Nets on a real-life black-box model, i.e. without the important prior knowledge provided by pre-trained weight on the same dataset as $F_v$, ResNet-34 and ResNet-101 showed similar performances.

Therefore, whether the $F_v$ is black-box or white-box, an optimized structure of $F_a$ needs to be searched. The best strategy for the attacker might be to adopt an AutoML approach, search for the optimal hyperparameters according to his budget and prior knowledge, and continuously adjust them with the query results he obtains. 

\subsubsection{Optimizer}
\label{subsec:opt}
Besides model structure, another typical hyperparameter that needs to be adaptively configured is the optimizer for training $F_a$. One significant difference for optimizer choice compared to the typical machine learning is that, overfitting the training accuracy might not be bad for testing accuracy. In this paper, $F_a$ are trained for a large number of epochs with little learning rate decay, which by common sense will lead to overfitting, but the test accuracy is observed to be continuously improving. We believe that in the scenarios where the adversary does not have a good prior knowledge about the dataset, his best strategy is to overfit to the soft labels. In addition, a typical training dataset has overall ``trending'' for each class, whereas the distribution of $D_t$ is more dynamic during the extraction process. One interesting observation we obtained in our experiments is that Adam optimizer always leads to a significantly worse $F_a$ than SGD does. Thus, optimizers validated in common machine learning scenarios may not be directly applicable in model extraction.

\begin{table}[t]
\caption{Results of model extraction on FashionMNIST with Specific Classes Excluded}\label{tab:sl}
\begin{center}
\begin{tabular}{|c|c|c|c|}
\hline
$\boldsymbol{D_a}$ & \textbf{Excluded $\boldsymbol{K}$ Class(es)}$^{\mathrm{a}}$&$\boldsymbol{{R}_{EC}}$~\textbf{(\%)} & $\boldsymbol{{P}_{EC}}$~\textbf{(\%)}\\
\hline
FMNIST-1 & \multirow{2}{*}{9} & 92.9 & 97.8\\ \cline{1-1} \cline{3-4}
FMNIST-1S & &86.0 & 98.9\\ \cline{1-4}
FMNIST-2 & \multirow{2}{*}{1,9} & 93.2 & 98.9\\ \cline{1-1} \cline{3-4}
FMNIST-2S & &79.6 & 99.1\\ \cline{1-4}
FMNIST-3 & \multirow{2}{*}{0,1,9}& 90.6 & 93.2\\ \cline{1-1} \cline{3-4}
FMNIST-3S & &75.5 & 96.0\\ \cline{1-4}
FMNIST-8 & \multirow{2}{*}{0,1,2,3,4,5,7,9} &68.7 & 84.4 \\ \cline{1-1} \cline{3-4}
FMNIST-8S &  & 16.9 & 64.6\\ 
\hline
\multicolumn{4}{l}{$^{\mathrm{a}}$Mapping from number to class name follows the official data source.} \\
\end{tabular}
\end{center}
\end{table}

\subsection{Query Utility}
\label{subsec:Information}
In experiments shown in Table \ref{tab:structure}, we notice that $F_a$ can learn features by querying with white noise, which is meaningless to humans and contains no perceivable feature related to the task. To further explore this phenomenon, we conduct a series of experiments of extracting $F_v$ on FashionMNIST with varying $D_a$:
\begin{itemize}
    \item FMNIST-$K$: the FashionMNIST training dataset, excluding all data from the specified $K$ class(es).
    \item FMNIST-$K$-S: the FashionMNIST training dataset, excluding all samples to which $F_v$ assigns a confidence score larger than 10\% on the specified $K$ class(es)".
\end{itemize}

All available samples in the $D_a$ are queried to the victim exactly once, with no data argumentation applied. We consider the performance of $F_a$ on the ``excluded class(es)", and evaluate the prediction with two metrics, namely:
\begin{enumerate}
    \item Recall ${R}_{EC}$: defined as the ratio of correct predicted labels in excluded classes and the total number of images in these classes
    \item Precision ${P}_{EC}$: defined as the ratio of correct predicted labels in excluded classes and the total number of prediction in these classes
\end{enumerate}
We notice that despite images of certain classes are missing in $D_a$, $F_a$ is still able to classify them with high precision, even in extreme cases where all sampled data obtained have low confidence in the target class. Interestingly, the precision ${P}_{EC}$ is higher than the recall ${R}_{EC}$. It seems $F_a$ is more conservative in ``excluded classes''.

This experiment shows that an adversary might be able to learn without prior knowledge of the target class(es). Additionally, as a further investigation,  we query the victim with MNIST \cite{lecun-mnisthandwrittendigit-2010} dataset as $D_a$, which are typical out-of-distribution data in this case, and $F_a$ achieved 66.7\% test accuracy on the FashionMNIST test dataset. It seems in these cases, misclassified queries are no longer adversarial dirty data and, instead, they play a positive role in the extraction. Meanwhile, soft label with relative growth and drop across classes makes the learning from 'low confidence' samples possible and facilitates the extraction when lacking input data. We believe that explains why in Table \ref{tab:structure}, the overall extraction result degenerates with a more accurate victim. Similar observations were also made by \cite{yang2018knowledge, cho2019efficacy} in the context of knowledge distillation.

\begin{figure*}[t]
  \includegraphics[width=\textwidth]{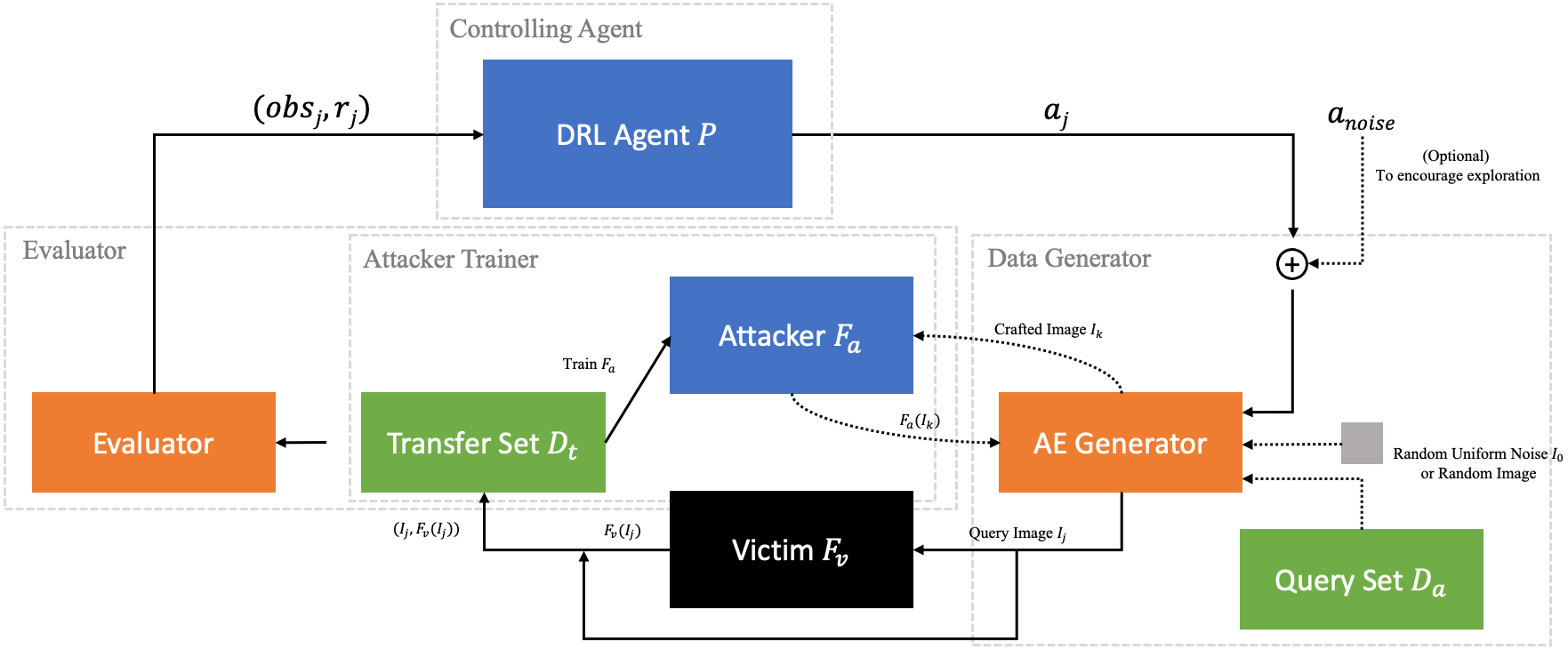}
  \caption{Structure of the proposed framework. Modules are indicated by grey dot-lined boxes with names on the top-left corner. Detailed components within each module are from Algorithm \ref{alg:algorithm}. Blue box ($P$, $F_a$) denotes trainable network. Green box ($D_t$, $D_a$) denotes data storage. Orange box (Evaluator, AE Generator) denotes fixed function. Black box ($F_v$) denotes black-box component.} \label{fig:struc}
\end{figure*}

\begin{algorithm}[tb]
\caption{DRL Guided Model Extraction with i-FGSM: An Implementation of the Proposed Framework} \label{alg:algorithm}
\textbf{Input}: Victim Model $F_v$, Data Pool $D_a$, Attacker Model $F_a$, \\
\textbf{Parameter}: Query Budget $B$, Evaluation Scope $X$, FGSM Iteration $M$\\
\textbf{Output}: Trained $F_a$ that shares similar functionality with $F_v$

\begin{algorithmic}[1]
\STATE Initialize Controller $P$ and fill $obs_0$ with zeros
\STATE Let $j = 0$
\WHILE{$j \leq B$}
\STATE Feed forward $obs_j$ into $P$ to output a target probability distribution ${prob}_t$ and $\epsilon$
\IF {$D_a$ is not empty}
\STATE Randomly draw an image from $D_a$ as $I$
\ELSE
\STATE Generate a noise image with uniform random as $I$
\ENDIF
\STATE Apply i-FGSM with $\epsilon$ on $I$ for $M$ iterations, note the output as ${I}_{j}$
\STATE Store $({I}_{j}, F_v({I}_{j})$ pair into $D_t$
\STATE Train $F_a$ with last $X$ samples in $D_t$
\STATE Evaluate the distribution and the loss of $F_a$ on the last $X$ samples in $D_t$ with \eqref{form:std_p}, \eqref{form:train_loss}, and \eqref{form:range_p}. Set the evaluation result as $obs_j$
\STATE Calculate the reward $r_j$ with \eqref{form:rwd}
\STATE Update $P$ 
\ENDWHILE
\STATE \textbf{return} $F_a$
\end{algorithmic}
\end{algorithm}

\section{Method and Evaluation}
\label{sec:experiment}
With observations made in section \ref{subsec:Information}, on the top of \cite{orekondy2019knockoff}, we hereby present an abstract framework (Fig. \ref{fig:struc}) for model extraction analysis and give out a heuristic example of implementation (Algorithm \ref{alg:algorithm}) in the context of image classification.
\subsection{Framework}
The framework consists of four loose-coupling modules: a Data Generator that generates queries based on available resources and an instruction given by a Controlling Agent. The queries are then sent to the victim model, and an attacker trainer will train the model based on the queries and results. An Evaluator is deployed to observe the extraction efficiency and provides feedback to the Controlling Agent. The detailed functionality and design choices of each module are as follows:
\subsubsection{Controlling Agent}
Controlling Agent leads the Active Learning process. Based on the current extraction state, it issues instructions for the environment to execute. In this paper, we use DDPG \cite{lillicrap2015continuous} to output desired soft-label $p_t$ and FGSM parameter $\epsilon$ for the next query as an example. However, it is also compatible with other continuous action space DRL algorithms such as TRPO \cite{DBLP:journals/corr/SchulmanLMJA15}. Classic controlling algorithms such as MPC \cite{amos2018differentiable} may also be adopted in this scenario.

\subsubsection{Data Generator}
Data Generator is responsible for providing query images to the black-box victim model that full fills the instruction given by Controlling Agent. In this paper, we randomly select or generate base images, and adopt i-FGSM \cite{kurakin1607adversarial}, utilizing the transferability of AE, to enrich non-robust features \cite{ilyas2019adversarial} covered in $D_t$. Given the flexibility of this module, we believe that methods like prioritized base image selection or stronger white-box AE attack could be applied to further enhance the performance. We cater them for future research.

\subsubsection{Attacker Trainer}
In our proposed framework, the attacker model is trained during the sampling process, in order to timely reflect the state of extraction. Considering the specialty of query data distribution (section \ref{subsec:struc}), we use SGD as our optimizer. After each query image is made to the victim model, we train the attacker model on $X$ most recently collected samples to reduce the time taken by the sampling process. However, as a reflection of the stateful model stealing defense \cite{10.1145/3274694.3274740}, we believe sampling time is not a critical factor to be considered for the whole extraction process, and better results could be achieved by a more sufficiently trained attacker.

\subsubsection{Evaluator}
In deep reinforcement learning, feedback from the environment to the agent is critical, which consists of two parts: observation $obs$ and reward $r$. $obs$ reflects the state, or partial-state, of the current environment, and $r$ reflects how good did the agent do. Thus, Evaluator module gathers critical information about the current extraction process, then award or penalize the Controlling Agent. In Algorithm \ref{alg:algorithm}, we limit the evaluation scope to the latest $X$ elements in the transfer set. With $D_{X,j}$ denoting the newly added $X$ elements into the training set, i.e. $D_{X,j} = (D_{t,j}\setminus D_{t,j-X})$, the $j$-th $obs$ contains following information:
\begin{itemize}
    \item $F_a$ output on $I_{j}$ in probability: $F_a(I_j)$
    \item $F_v$ output on $I_{j}$ in probability: $F_v(I_j)$
    \item The mean $\bar{m}_{j}$ \eqref{form:mean_p} and standard deviation $\sigma_j$ \eqref{form:std_p} of the (soft) labels for each class in $D_{X,j}$.
    
\begin{equation}
    \bar{m}_{j} = \frac{1}{j}\sum_{x\in D_{X,j}}{F_v (x)}\label{form:mean_p}
\end{equation}%

\begin{equation}
    d_{j} = \sqrt{\frac{1}{j}\sum_{x\in D_{X,j}}(F_v(x)- \bar{m}_{j})^2}\label{form:std_p}
\end{equation}% 
    \item Mean Cross-entropy Loss of $F_a$ on each class \eqref{form:train_loss} with $D_{t,j}$. 
\begin{equation}
    \bar{L}_{j} = \frac{1}{j}\sum_{x \in D_{X,j}} L_{CE}(F_a(x), F_v(x))\label{form:train_loss}
\end{equation}%

\end{itemize}
where $L_{CE}(F_a(x), F_v(x))$ is the Cross-entropy between $F_a(x)$ and $F_v(x)$. Given $obs$, the agent will decide an action that attempts to maximize the accumulated reward, calculated by a reward function. The reward function shall assess whether the changes of the transfer set will lead to a better attacker model,  with respect to the newly added elements. The observation made in section \ref{subsec:Information} shows that we shall encourage diversity in classes, and discourage similar samples to existing ones in $D_t$. Let $\sigma(m)$ denote the standard deviation of the given vector across all elements, we use the following equations for reward in algorithm \ref{alg:algorithm}:

\begin{itemize}
 \item the Cross-Entropy Loss of $F_a$ for the last $X$ elements. Note that a higher loss indicates that the samples are not similar to existing ones, and thus shall be encouraged, in order to explore the input space. That is:
\begin{equation}
    r_{ce}=\left\{
    \begin{array}{rcl}
    1 & & {\bar{L}_{j} > \bar{L}_{j-1}}\\
    -1 & & {\bar{L}_{j} \leq \bar{L}_{j-1}}\\
    \end{array} \right. \label{form:r_ce}
\end{equation}%
 \item the diversity of samples. This is measured by both the range and standard deviation of $\bar{m}_{j}$ , as well as encouraged by the minimum standard derivation of probabilities from each class:
\begin{equation}
    R_j = \max \bar{m}_{j} - \min \bar{m}_{j}\label{form:range_p}
\end{equation}%
 \begin{equation}
    r_{range}=\left\{
    \begin{array}{rcl}
    1 & & {R_j < R_{j-1}}\\
    -1 & & {R_j \geq R_{j-1}}\\
    \end{array} \right. \label{form:r_range}
\end{equation}%

\begin{equation}
    r_{std}= \min d_{j} \label{form:r_std}
\end{equation}%

\begin{equation}
    r_{avg}=\left\{
    \begin{array}{rcl}
    1 & & {\sigma(\bar{m}_j) < \sigma(\bar{m}_j})\\
    -1 & & {\sigma(\bar{m}_j) \geq \sigma(\bar{m}_j})\\
    \end{array} \right.\label{form:r_avg}
\end{equation}%
 
\end{itemize}
The final reward is the weighted sum of the above factors:

\begin{equation}
  \begin{aligned}
      r = & \lambda_0 * r_{ce} + \lambda_1 * r_{rang} \\+ & \lambda_2 * r_{std} + \lambda_3 * r_{avg} \label{form:rwd}
 \end{aligned}
\end{equation}%

\subsection{Benchmarking}
We now evaluate the performance of the proposed Algorithm \ref{alg:algorithm} by attacking a $F_v$ trained with the FashionMNIST dataset. In order to simulate the scenarios where the attacker has limited prior knowledge about the victim, we strictly limit the amount of data contained in $D_a$. Additionally, we test one variant, where the value of $r_{ce}$ \eqref{form:r_ce}, $r_{range}$ \eqref{form:r_range} and $r_{avg}$ \eqref{form:r_avg} are computed directly instead of discretely based on relative change. To validate the effectiveness of DRL and i-FGSM, we also included results of the following two baselines, modified from Algorithm \ref{alg:algorithm} but within our proposed framework:
\begin{enumerate}
    \item Random Uniform Noise as Data Generator
    \label{baseline1}
    \item Random Controller as Controlling Agent
    \label{baseline2}
\end{enumerate}
Note that \cite{orekondy2019knockoff} is not included in the benchmarking baselines, for it is not designed for zero or limited unlabeled data conditions. To our best knowledge, we are the first to attempt data-free stealing in the context of computer vision.

To demonstrate the adaptiveness of our method, in this experiment, hyperparameters are set intuitively and are not specifically tuned. All $\lambda$ in the reward function are set to be 1. The query budget $B$ is set to $10^4$, which is one-sixth of the FashionMNIST training set size. To encourage exploration, we apply a Gaussian noise $n \sim N(0, 0.1)$ to the DRL action for the first $2 * 10^3$ steps. To balance the trade-off between attacker training and experiment run-time, we set the evaluation scope $X$ to be 640. Meanwhile, guided by the findings in section \ref{subsec:opt}, we additionally train each $F_a$ for 1000 epochs with $D_a$ after the sampling process for better information utilization.

The result (Table \ref{tab:drl}) shows that in both data-free and limited-data situation, Algorithm \ref{alg:algorithm} and its variant outperform the two baseline methods. Notably, in ``Limited Data" situation, where $D_a$ contains 20 randomly drawn images from a pool of $6*10^{4}$, our method achieves a more stable result than the baseline ones, which demonstrates the efficiency of DRL and adaptiveness of our proposed framework. Moreover, Algorithm \ref{alg:algorithm} saves the human effort of collecting or organizing a vast amount of candidate data, and instead, makes use of non-robust features which is easier to be crafted. From another aspect, this benchmarking experiment also suggests that an attacker does not need to be well-prepared to attack a victim, deepening concerns of protecting commercial models against model extraction attacks as crime cost is significantly lower.

\begin{table}[t]
\caption{Benchmarking Result of Algorithm 1 on FashionMNIST}\label{tab:drl}
\begin{center}
\begin{tabular}{|c|c|c|}
\hline
\multirow{2}{*}{\textbf{Method}}&\multicolumn{2}{|c|}{$\boldsymbol{ACC}^{\mathrm{a}}$~\textbf{(\%)}} \\
\cline{2-3} 
& \textbf{\textit{Data Free}}$^{\mathrm{b}}$& \textbf{\textit{Limited Data}}$^{\mathrm{c}}$ \\
\hline
Algorithm \ref{alg:algorithm} & \textbf{54.7$\pm$1.4} & 75.2$\pm$1.8\\ \cline{1-3}
Algorithm \ref{alg:algorithm} (Variant) & 52.5$\pm$1.9 & \textbf{75.4$\pm$0.9}\\ \cline{1-3}
Baseline 1 & 32.9$\pm$8.0 & 62.2$\pm$5.3\\ \cline{1-3}
Baseline 2 & 44.2$\pm$1.6 & 72.7$\pm$2.5\\ \cline{1-3}
\hline
\multicolumn{3}{l}{$^{\mathrm{a}}$$ACC$: Observed best top-1 test accuracy.} \\
\multicolumn{3}{l}{$^{\mathrm{b}}$Data Free: $D_a$ contains no data.} \\
\multicolumn{3}{l}{$^{\mathrm{c}}$Limited Data: $D_a$ contains 20 unlabeled in-distribution images.} \\
\end{tabular}
\end{center}
\end{table}

\section{Conclusion and Future Work}
\label{sec:conclusion}
In this paper, we study the model extraction problem when the attacker does not have much prior knowledge about in-distribution and lack pre-trained networks to start with. We empirically find that the statements made by the current literature fail to extend to such stricter scenarios. Meanwhile, we verified that, although the difficulty of extraction is significantly higher, an attacker can still learn information with out-of-distribution data. We study the different factors that contribute to the model extraction and put together these factors into a framework that can be trained for better configurations. We conduct experiments to demonstrate the effectiveness of the algorithm to extend the existing FGSM method with little prior knowledge of the victim model. Thus, when companies that provide AIaaS shall be aware that it is vulnerable when the confidentiality of their models rely only on the secrecy of their data.

For simplicity of illustration, factors such as attacker model structure are not covered by our given example (Algorithm \ref{alg:algorithm}). Whether techniques like neural architecture search (NAS) \cite{ren2020comprehensive} in the current literature could be applied to improve the current implementation is still left for further research. Moreover, Algorithm \ref{alg:algorithm} adopts DDPG, which is a model-free reinforcement learning algorithm. In our benchmarking experiments, the DRL agent is trained from scratch as the sampling carries out. We leave techniques to reduce the DRL converging steps like model-based reinforcement learning \cite{bansal2017mbmf} or meta reinforcement learning \cite{duan2016rl} in future work.

\bibliographystyle{IEEEtran}
\bibliography{IEEEabrv, ref}

\end{document}